\documentclass{article}
\usepackage{ijcai24}

\usepackage{times}
\usepackage{soul}
\usepackage{url}
\usepackage[hidelinks]{hyperref}
\usepackage[utf8]{inputenc}
\usepackage[small]{caption}
\usepackage{graphicx}
\usepackage{amsmath}
\usepackage{amsthm}
\usepackage{booktabs}
\usepackage{algorithm}
\usepackage{algorithmic}
\usepackage[switch]{lineno}
\usepackage{multirow}

\title{Paintings and Drawings Aesthetics Assessment with Rich Attributes\\ for Various Artistic Categories}

\author{
    Xin~Jin$^1$\and
    Qianqian~Qiao$^1$\and
    Yi~Lu$^2$\and
    Huaye~Wang$^1$\and
    Shan~Gao$^3$\and
    Heng~Huang$^4$\footnote{Corresponding author: Heng Huang(hecate@mail.ustc.edu.cn)}\and
    Guangdong~Li$^5$
    \affiliations
        $^1$Beijing Electronic Science and Technology Institute\\
        $^2$Central Academy of Fine Arts\\
        $^3$Beijing University of Technology\\
        $^4$University of Science and Technology of China\\
        $^5$Beijing Institute of Fashion Technology\\
}

\begin{document}
\maketitle
\begin{abstract}
Image aesthetic evaluation is a highly prominent research domain in the field of computer vision. In recent years, there has been a proliferation of datasets and corresponding evaluation methodologies for assessing the aesthetic quality of photographic works, leading to the establishment of a relatively mature research environment. However, in contrast to the extensive research in photographic aesthetics, the field of aesthetic evaluation for paintings and Drawings has seen limited attention until the introduction of the BAID dataset in March 2023. This dataset solely comprises overall scores for high-quality artistic images. 
Our research marks the pioneering introduction of a multi-attribute, multi-category dataset specifically tailored to the field of painting: Aesthetics of Paintings and Drawings Dataset (APDD). 
The construction of APDD received active participation from 28 professional artists worldwide, along with dozens of students specializing in the field of art. This dataset encompasses 24 distinct artistic categories and 10 different aesthetic attributes. Each image in APDD has been evaluated by six professionally trained experts in the field of art, including assessments for both total aesthetic scores and aesthetic attribute scores. The final APDD dataset comprises a total of 4985 images, with an annotation count exceeding 31100 entries.
Concurrently, we propose an innovative approach: Art Assessment Network for Specific Painting Styles (AANSPS), designed for the assessment of aesthetic attributes in mixed-attribute art datasets. Through this research, our goal is to catalyze advancements in the field of aesthetic evaluation for paintings and drawings, while enriching the available resources and methodologies for its further development and application.

\end{abstract}

\section{Introduction}

Computational aesthetics~\cite{datta2006studying} aims to enable computers and robots to recognize, generate, and create beauty. In related research, computational visual aesthetics ~\cite{brachmann2017computational} primarily involves training large datasets to acquire neural network models, enabling the models to provide evaluations of aesthetic quality. Consequently, the construction of benchmark datasets for Image Aesthetic Quality Assessment (IAQA) has become a crucial prerequisite for advancing research in this direction. However, existing datasets predominantly focus on total aesthetic scores of images, with limited exploration in the study of image categories and aesthetic attributes. Moreover, the majority of existing datasets are concentrated in the field of photo, with sparse representation in the field of artistic images.
\begin{figure}[htbp]
  \centering
    \includegraphics[width=0.8\linewidth]{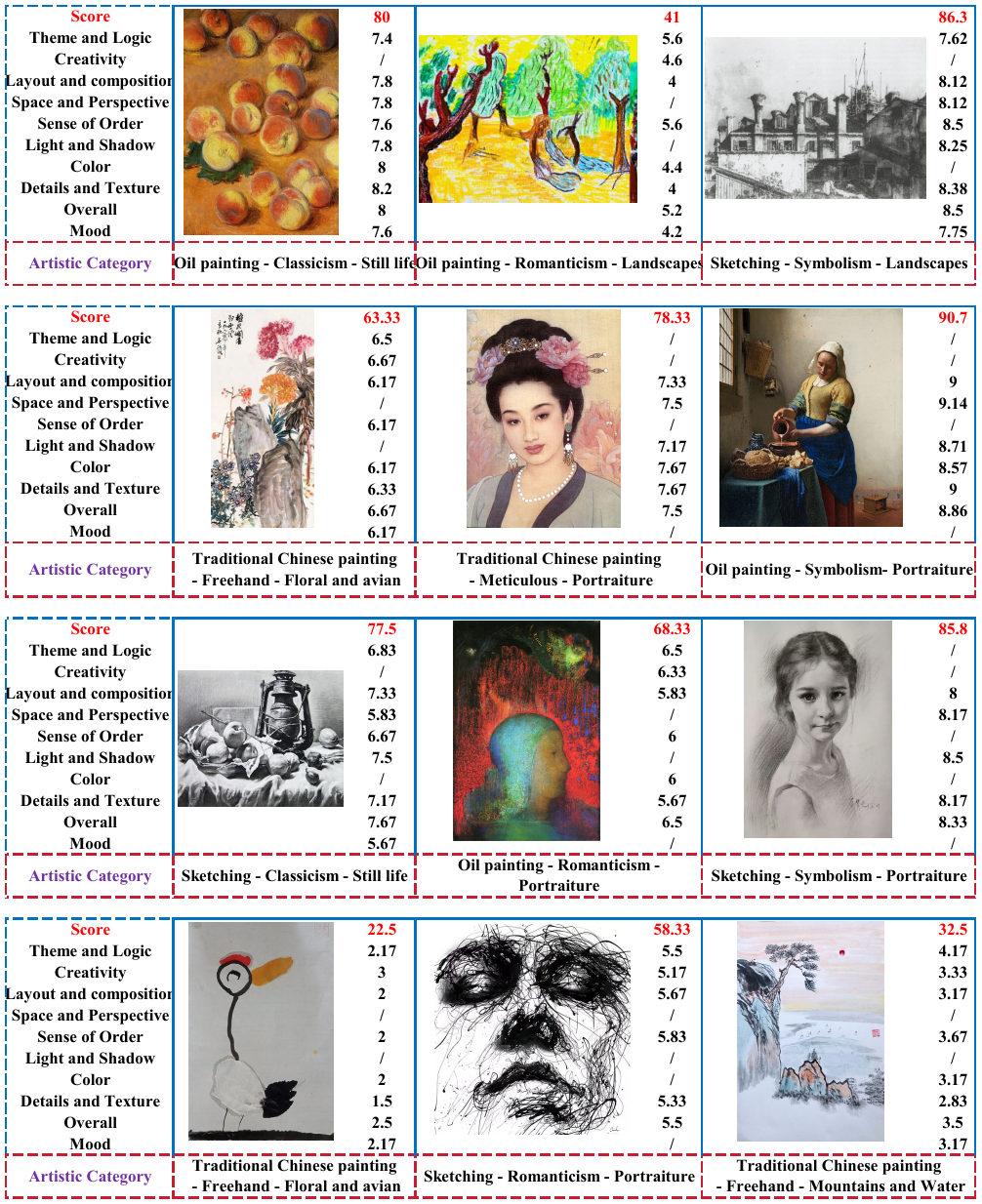}
   \caption{Samples from the APDD dataset. APDD covers 24 artistic categories and 10 aesthetic attributes. Different artistic categories correspond to different sets of attributes.}
   \label{IAQA Task & Dataset}
\end{figure}

\begin{table*}[htbp]
    \centering
    \begin{tabular}{cccccc}
        \hline
        \textbf{\multirow{2}{*}{Field}} & \textbf{\multirow{2}{*}{Dataset}} & \textbf{Number of} & \textbf{Number of} & \textbf{Number of} & \textbf{\multirow{2}{*}{IAQA Task Types}} \\
        & & \textbf{images} & \textbf{attributes} & \textbf{categories}  & \\
        \hline
        \multirow{6}{*}{Photo Field} & AVA ~\cite{murray2012ava}  & 255,530  & - & - & Aesthetics Score\\
        & AADB ~\cite{kong2016photo}  & 10,000  & 11 & - &  Aesthetic Attributes\\
         & PCCD ~\cite{chang2017aesthetic}  & 4,235 & 7 & - & Aesthetic Captions  \\
        & AROD ~\cite{schwarz2018will}  & 380,000 & - & - & Aesthetics Score\\
        & AVA-Reviews ~\cite{wang2019neural}  & 52,118 & - & - & Aesthetic Captions\\
        & EVA ~\cite{kang2020eva}  & 4,070  & 4 & 6 &  Aesthetic Attributes\\
        \hline
        Art Field & BAID ~\cite{yi2023towards}  & 60,337 & - & - & Aesthetics Score\\
        \hline
        
        \multirow{5}{*}{Art Field}& VAPS ~\cite{fekete2022vienna}  & 999& 5 & 5 &  \multirow{5}{*}{Aesthetic Attributes}\\
         & JenAesthetics ~\cite{amirshahi2015jenaesthetics}  & 1,268 & 5 & 16 &\\
        & JenAesthetics$\beta$ ~\cite{amirshahi2016color}  & 281 & 1 (beauty) & 16 & \\
        & MART ~\cite{yanulevskaya2012eye}  & 500 & 1 (emotion) & - &  \\
        & \textbf{APDD (Ours)}  & \textbf{4,985} & \textbf{10} & \textbf{24} &\\
        \hline
    \end{tabular}
  \caption{A comparison between the APDD dataset and existing image datasets.}
  \label{tab: Comparison of datasets}		
\end{table*}

Existing artistic image aesthetic datasets (such as BAID ~\cite{yi2023towards}, VAPS ~\cite{fekete2022vienna}, JenAesthetics ~\cite{amirshahi2015jenaesthetics}) suffer from limitations in both image quantity and the absence of annotations for aesthetic attributes. This deficiency results in a weak interpretability of overall aesthetic scores, rendering them less persuasive. Additionally, these images predominantly represent either iconic works from the history of painting art or high-quality competition entries, lacking a representation of medium to low-quality painting works. Furthermore, these datasets predominantly focus on oil painting as the sole painting form, neglecting the diverse and extensive nature of painting as an art form, which encompasses various classifications. A singular dataset composed solely of high-quality oil paintings lacks diversity, making it unsuitable for multi-task learning and challenging to address the complexities of real-world scenarios.

To address the deficiencies in existing artistic image datasets, such as insufficient aesthetic attributes, uneven aesthetic quality, and limited artistic categories, we leveraged the knowledge and professional expertise of 28 professional artists in the field of painting. Successfully, we constructed the first-ever multi-attribute, multi-category dataset specifically tailored for the field of painting: the Aesthetics of Paintings and Drawings Dataset (APDD). APDD is structured into 24 distinct artistic categories based on different painting categories, artistic styles, and subject matter. Additionally, we selected 10 aesthetic attributes for APDD, including theme and logic, creativity, layout and composition, space and perspective, sense of order, light and shadow, color, detail and texture, overall, and mood. Based on the characteristics of different artistic categories, we selectively chose distinct sets of aesthetic attributes tailored to each artistic category. Additionally, we introduce a paintings and drawings assessment network named Art Assessment Network for Specific Painting Styles (AANSPS), designed to evaluate aesthetic attributes in mixed-attribute painting datasets.

In conclusion, the main contributions of this paper are as follows:

Firstly, we propose a clear framework for considering aesthetic components in artistic images, providing a detailed categorization of artistic categories in the field of painting along with their corresponding aesthetic attributes. We also establish scoring criteria for these attributes within different artistic categories.

Secondly, we address the gap in the art field for the lack of a multi-attribute, multi-category painting dataset, introducing the Aesthetics of Paintings and Drawings Dataset (APDD) for the first time.

Lastly, to assess the total aesthetic scores and aesthetic attribute scores of paintings, we propose a painting image evaluation network called AANSPS. We evaluate advanced image aesthetic assessment methods and AANSPS on the APDD dataset. Our model achieves satisfactory results across all metrics, providing clear evidence of the effectiveness of our approach.

\section{Related Work}
\subsection{Image Aesthetic Assessment Datasets}

The AVA dataset ~\cite{murray2012ava} pioneered the use of large-scale images for aesthetic analysis, thereby advancing research in computational aesthetics. In 2015, the JenAesthetics dataset ~\cite{amirshahi2015jenaesthetics} was introduced, containing 1,628 colored oil paintings exhibited in museums, with each image annotated for five attribute scores. The AADB dataset ~\cite{kong2016photo}, released in 2016 , consists of 10,000 photographic images annotated by five individuals. Released in 2017, the PCCD dataset ~\cite{chang2017aesthetic} encompasses 4,307 photographic images with annotations providing language comments and aesthetic scores for seven aesthetic attributes. Subsequently, the 2018 release of AROD ~\cite{schwarz2018will} and the 2019 release of AVA-Reviews ~\cite{wang2019neural} expanded the scale of data for image aesthetic scorings, aesthetic classification, and language comments. In 2022, the VAPS dataset ~\cite{fekete2022vienna} featured 999 representative works from the history of painting art. In 2023, the BAID dataset ~\cite{yi2023towards} was introduced, comprising 60,337 high-quality artistic images, each associated with an overall aesthetic score.

Among all these datasets, the majority of images are concentrated in the field of photo, with relatively fewer artistic images. Furthermore, most datasets do not adequately classify images, and the attribute types in existing artistic image datasets cannot comprehensively reflect the fundamental characteristics of paintings. A detailed comparison of our APDD with existing datasets is provided in Table~\ref{tab: Comparison of datasets}.

\subsection{Image Aesthetic Assessment Models}

Due to the powerful feature learning capability of deep networks, there has been an emergence of methods in recent years for predicting aesthetic attributes of images. Kong et al. (2016) ~\cite{kong2016photo} introduced an attribute-adaptive deep convolutional neural network for aesthetic score prediction, providing evaluations of aesthetic attributes simultaneously with aesthetic score prediction. Malu et al. (2017) ~\cite{malu2017learning} jointly learned aesthetic scores and aesthetic attributes using a deep convolutional network with merging layers. Building upon this, Pan et al. (2019) ~\cite{pan2019image} explored the inherent joint distribution of real aesthetic scores and attributes through adversarial learning to further enhance the prediction accuracy of aesthetic attributes and scores. Li et al. (2022) ~\cite{leida2022} proposed a deep multi-task convolutional neural network (MTCNN) model that leverages scene information to assist in predicting aesthetic attributes of images. Jin et al. (2023) ~\cite{jin2023aesthetic} presented a method for image aesthetic attribute assessment that achieves aesthetic classification, overall score, and scores for three attributes: light, color, and composition.

The aforementioned image aesthetic attribute evaluation methods are all proposed based on photographic images. In the field of artistic images, the AANSPS network proposed in this paper is capable of simultaneously extracting the total aesthetic scores and aesthetic attribute scores.

\section{Aesthetics of Paintings and Drawings Dataset}

\subsection{Formation of Professional Artistic Team}
In order to ensure the professionalism and authority of the developed painting dataset, we established a team of painting experts. These team members bring to the table extensive aesthetic experience and educational backgrounds garnered from art institutions worldwide. Their invaluable assistance encompassed the selection of artistic categories, identification of aesthetic attributes, the addition of paintings, and the scoring and annotation of images within the dataset. Specifically, the team is composed of artists, educators, and master's degree candidates specializing in oil painting, sketching, and Chinese painting, spanning both domestic and international contexts. Stringent criteria were established for all team members, requiring: 1) possession of a bachelor's degree or higher; 2) a minimum of 7 years of art education experience to ensure a profound artistic background; 3) reception of professional education in oil painting, sketching, or Chinese painting, along with the ability to evaluate artworks based on scoring standards; 4) affiliation with a professional art institute to guarantee that team members originate from institutions with robust artistic education backgrounds; 5) involvement in experiences such as exhibiting works in art exhibitions or publishing articles in professional journals.

Following these criteria, we have successfully assembled a team of specialized dataset builders, comprising 28 professional artists and 24 students with high academic qualifications. The team was segmented into a team leader (1 person), an oil painting group (19 people), a sketching group (19 people), and a Chinese painting group (13 people), with the most artistically adept member serving as the leader in each group. This team composition not only ensures a wealth of professional knowledge and extensive experience but also provides a more nuanced artistic perspective for scoring and annotation activities.

\subsection{Artistic Categories}

According to different painting category (oil painting, sketching, and traditional Chinese painting), artistic styles (symbolism, classicism, romanticism, meticulous, and freehand), and subject matter (landscapes, still life, portraiture, floral and avian, mountains and water), we have categorized the APDD dataset into 24 distinct artistic categories.

\begin{figure}[htbp]
  \centering
    \includegraphics[width=0.4\linewidth]{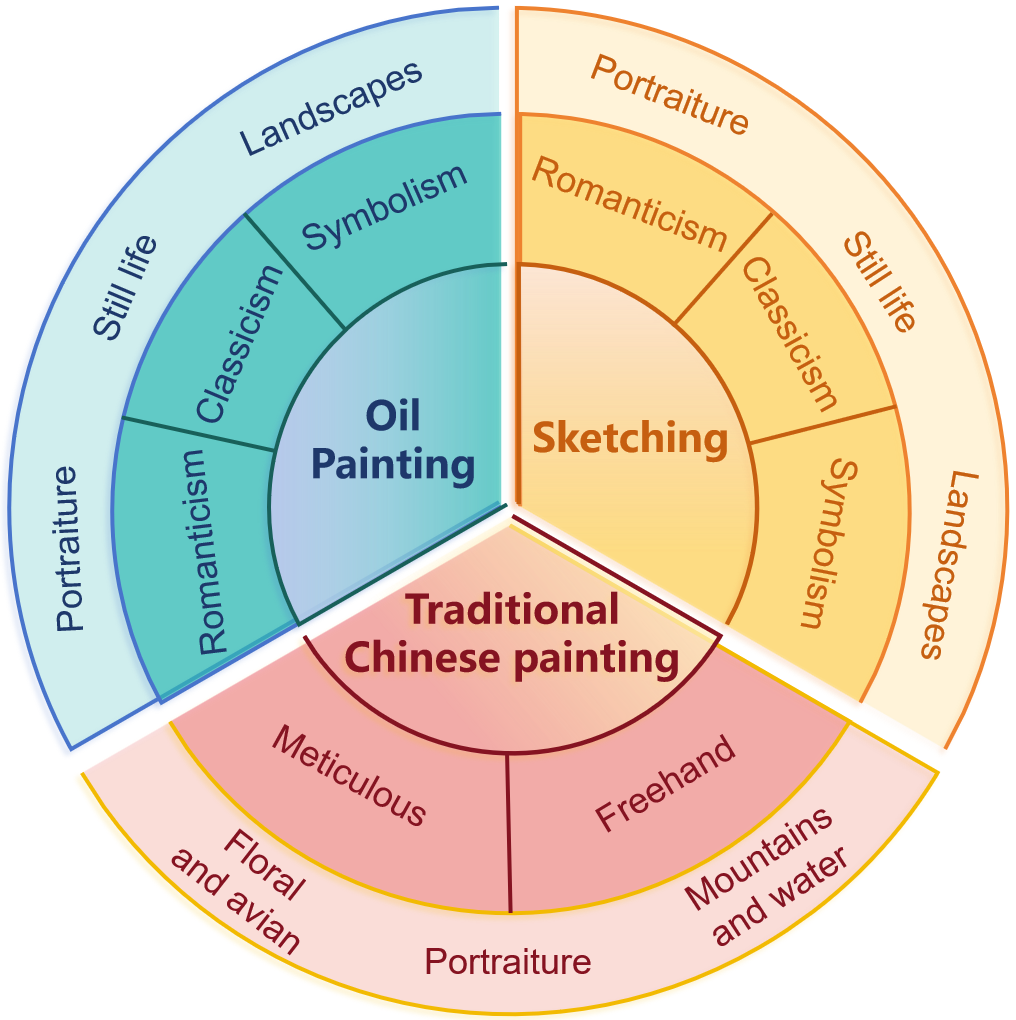}
   \caption{24 Artistic Categories in the APDD Dataset.}
   \label{fig:Artistic Categories2}
\end{figure}

Painting encompasses various types, each distinguished by its unique tools, materials, and techniques. These include oil paintings, sketches, ink drawings, traditional Chinese painting, prints, watercolors, gouache, murals, and other forms~\cite{IntroductiontoArt}. Other types of paintings include illustrations, cartoons, abstracts, children's paintings, collages, airbrushes, and digital panels. After in-depth research and comparison, we have chosen sketching, oil painting, and traditional Chinese painting as the fundamental categories of painting types, owing to their long-standing historical significance, widespread dissemination, and popularity among the general public.

Concerning the classification of artistic styles, we have broadly divided them into two main categories: Western art styles represented by sketching and oil painting, and Eastern art styles represented by traditional Chinese painting.

In terms of Western artistic styles, the classification is primarily based on the categories proposed by the eminent 18th-century German philosopher and aesthetician, Hegel. According to his theory, artistic styles represent different stages of idealized development, encompassing primitive symbolism, classical art (such as Greco-Roman classical art), and modern romanticism~\cite{PhilosophyofFineArts}. Symbolism (primitive art) is characterized by a dominance of material elements, classicism (Greco-Roman) harmonizes material and spiritual factors, while romanticism (modern art) emphasizes the predominance of spiritual elements over material aspects.

In classifying Eastern art styles, the categorization is grounded in the techniques employed in traditional Chinese painting. Traditional Chinese painting techniques are primarily delineated into meticulous painting and freehand painting. Meticulous painting emphasizes exacting and precise techniques for representing objects, while freehand painting places greater emphasis on brushwork to convey the artist's emotions and artistic expression, underscoring the unity of form and spirit. 

From the perspective of subject matter, painting can be categorized into various genres, including portraiture, landscape, still life, wildlife, historical themes, and genre painting. For sketching and oil painting, we have selected the three most common subject matter categories: portraiture, landscapes, and still life, as the basis for classification. Traditional Chinese painting can be broadly categorized into three major genres: portraiture, mountains and water painting, and floral and avian painting.

\subsection{Aesthetic Attributes}

With the assistance of the professional artistic team and guidance from art instructional materials~\cite{Fundamentalsofmold-making}~\cite{dodson1990keys}~\cite{edwards1997drawing}~\cite{gombrich1995story}, we have compiled the attributes for aesthetic scoring, with the specific process outlined as follows:

Summarizing the general thought process of artistic creators: 

\textit{Thinking (Creativity and Imagination) → Composition (Two-dimensional Plane) → Perspective (Three-dimensional Space) → Structure (Internal Composition) → Color (Emotional Atmosphere) → Texture (Detail).}

Summarizing how art observers conduct layered observations: 

\textit{Outline → Structure → Texture → Lighting.}

Referring to the grading criteria for the fundamental painting exam, summarizing the evaluators' scoring approach: 

\textit{Content (closely related to the theme) → Creativity (innovative design and creation) → Layout → Logic (consistent with artistic technique and theme, appropriate layout and color coordination) → Form → Color → Details → Overall.}

Based on this, we summarize 10 aesthetic attributes for art images, namely, theme and logic, creativity, layout and composition, space and perspective, sense of order, light and shadow, color, detail and texture, overall, and mood.

\textbf{Theme and logic} require that the central idea and main content to be expressed in the work conform to the theme ~\cite{ART}.

\textbf{Creativity} requires that the work possesses creativity and imagination, and that the design is unique and able to break the traditional rules.

\textbf{Layout and composition} refer to the overall formal and structural relationship of the picture and its visual effect. Layout is the underlying logic of composition, and composition is the appearance of layout.

\textbf{Space and perspective} indicate the layering of space and the contrast between distance and nearness. Perspective is the basic factor in painting to express the three-dimensionality of objects and create spatial effects.

\textbf{The sense of order} is one of the important manifestations of the elements of perfection, coordination and wholeness of visual form, the core of which is some kind of consistency existing between the elements ~\cite{AnalysisofBeauty}.

The change of \textbf{light and shadow} makes the picture more visually rhythmic and rhythmic.

\textbf{Color} can render emotional atmosphere, reasonable color matching can make the picture achieve better visual effect.

\textbf{Details and texture} require a high degree of completion of the picture, with specific and vivid details, fine texture, to enhance the sense of reality of the picture.

\textbf{The overall} requirements of the picture in the overall effect of coordination, clear theme ~\cite{PhilosophyofArt}.

\textbf{Mood} refers to the poetic space that presents the blending of scene and reality, and the rhythm of active life ~\cite{TheJourneyofBeauty}.

\subsection{Correspondence between Artistic Categories and Aesthetic Attributes}

\begin{figure}[htbp]
  \centering
    \includegraphics[width=1\linewidth]{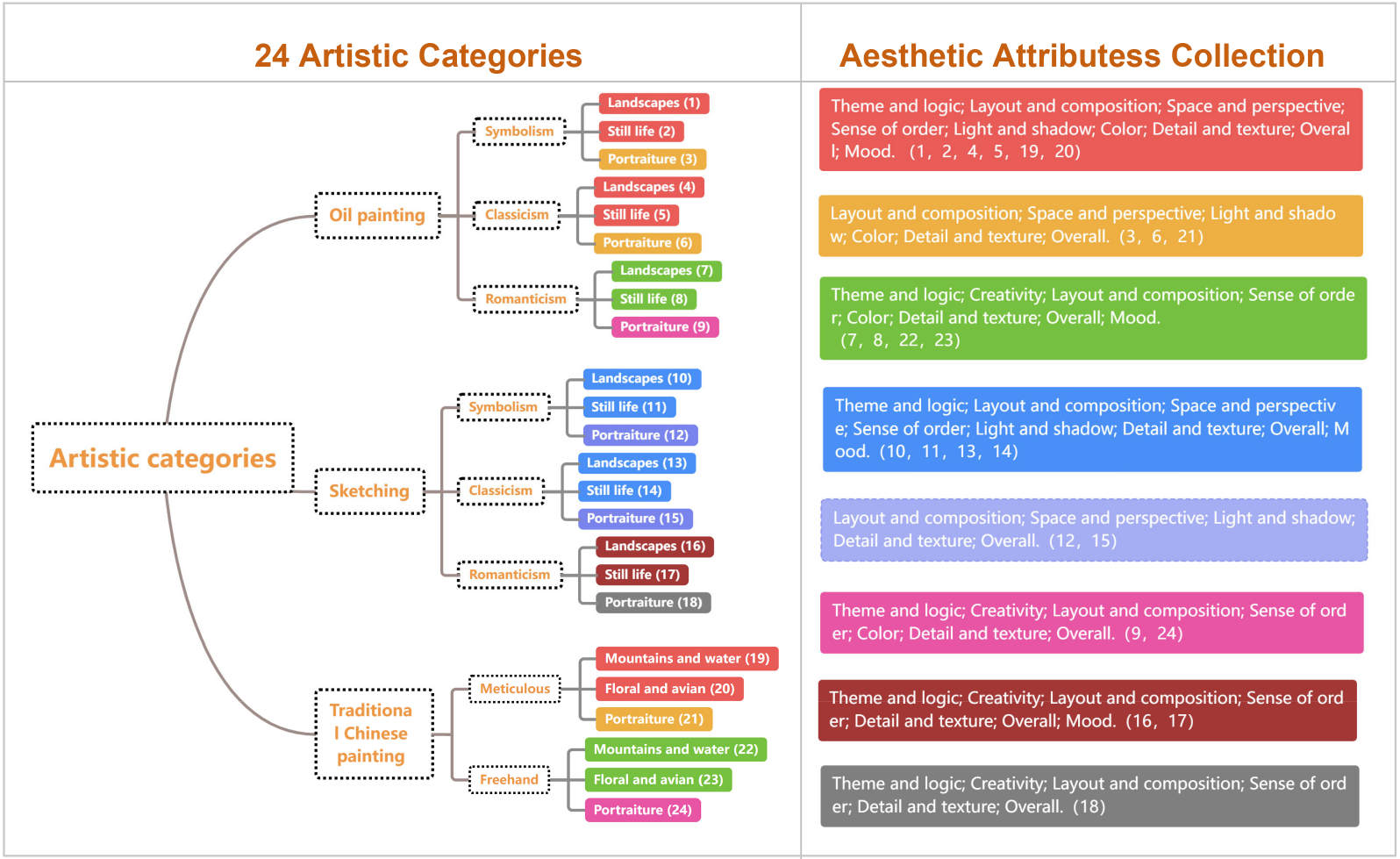}
   \caption{Correspondence between artistic categories and aesthetic attributes}
   \label{fig:oneco2}
\end{figure}

Just as sketching lack the attribute of color, not all artistic categories encompass the 10 aesthetic attributes proposed in this paper. The correspondence between artistic categories and aesthetic attributes is shown in Figure \ref{fig:oneco2}.

Artistic styles of oil painting include symbolism (1, 2, 3), classicism (4, 5, 6), and romanticism (7, 8, 9). Symbolism (1, 2, 3) can be understood as an objective reproduction, ignoring the attribute of "creativity", while other aesthetic attributes are more demanding. Classicism (4, 5, 6) can be understood as the unification of material and spiritual factors, i.e., "both form and spirit", and the attribute of "creativity" can be ignored. Among them, 3 and 6 are portraiture, in which "creativity" can be ignored along with "theme and logic", "sense of order" and "mood". Romanticism (7, 8, 9) can be understood as the spiritual factor exceeds the material factor, and can ignore the attributes of "space and perspective" and "light and shadow. Among them, 9 is portraiture, one can also ignore "mood".

Sketching is defined as the depiction of an object's outline, structure, light and shadow, and texture in monochromatic shades, so its "color" attribute can be ignored. Similar to oil painting, artistic styles of sketching are categorized as symbolism (10, 11, 12), classicism (13, 14, 15), and romanticism (16, 17, 18). 10, 11, 13, 14 can ignore "creativity". 12, 15 can ignore "theme and logic", "creativity", "sense of order", "mood". 16, 17 can ignore "space and perspective", "light and shadow". 18 can ignore "space and perspective", "light and shadow", and "mood".

Styles of Chinese painting are categorized as meticulous (19, 20, 21) and freehand (22, 23, 24). Meticulou can be understood as the use of neat and rigorous techniques to depict an object, and the attribute of "creativity" can be ignored in meticulous, while other aesthetic attributes are more demanding. Among them, 21 is portraiture, which can ignore "creativity" as well as "theme and logic", "sense of order" and "mood". The freehand can be interpreted as the author's subjective expression of things, and "space and perspective", "light and shadow" can be ignored. Among them, 24 is portraiture, and "mood" can also be ignored.

\subsection{Collection of Paintings and Drawings} 

To construct a comprehensive and diverse collection of paintings, we carefully selected several professional art websites and institutions as data sources to ensure the breadth and diversity of artistic images. The primary data sources chosen include ConceptArt\footnote{https://conceptartworld.com/}, DeviantArt\footnote{https://www.deviantart.com/}, WikiArt~\cite{phillips2011wiki}, the China Artists Association\footnote{https://caagov.com/}, and the National Art Museum of China\footnote{https://www.namoc.org/zgmsg/index.shtml}. The aesthetic quality of the paintings downloaded from these websites is concentrated in the intermediate and high ranges. To ensure the diversity and representativeness of the dataset, we introduced some artworks with lower aesthetic quality from the assignments of art students. Additionally, we adopted a strategy with a 3:1 ratio between works by professional artists and student assignments. This maintains the dominance of professional-level works in the dataset while also adequately considering the proportion of lower-scoring works.

In the end, we successfully collected 4,985 paintings, covering 24 categories, with each category containing a minimum of 200 images. This compilation includes works by renowned artists as well as student creations. The structure of this image dataset is designed to provide sufficient diversity and representativeness for subsequent scoring annotations.

\begin{table*}[htbp]
		
		\centering
		
  \resizebox{1.0\linewidth}{!}{%
		\begin{tabular}{cccccccc}
   
			\noalign{\smallskip}\hline\noalign{\smallskip}
			  & \textbf{Score} & \textbf{Theme and Logic} & \textbf{Creativity} & \textbf{Layout and composition} & \textbf{Space and Perspective}  \\
			
			\noalign{\smallskip}\hline\noalign{\smallskip}
			Number of annotations  & 31,100 & 24,595 & 11,702 & 31,100 & 19,398 \\
			
			Number of annotated persons  & 24 & 24 & 22 & 24 & 23  \\

                \noalign{\smallskip}\hline\noalign{\smallskip}
                 & \textbf{Sense of Order} & \textbf{Light and Shadow} & \textbf{Color} & \textbf{Details and Texture} & \textbf{Overall} & \textbf{Mood}\\

                \noalign{\smallskip}\hline\noalign{\smallskip}
                Number of annotations  & 24,595 & 19,398 & 18998 & 31,100 & 31,100  & 20,894\\

                Number of annotated persons &24 &23 & 22 & 24 & 24  & 24\\

			\noalign{\smallskip}\hline
		\end{tabular}
  }
    \caption{Annotation information statistics for the APDD dataset.}
    \label{tab: 2}
	\end{table*}

\subsection{Image Annotation Process} 

To ensure the effectiveness and smoothness of the annotation process, we specifically designed an online scoring system for the APDD dataset. In the development of the scoring system, we utilized VUE3 and the open-source framework Ruoyi for the frontend, while the backend was built on the Spring Boot framework. MySQL and Redis were chosen for the database. To operate the system more efficiently, we deployed it on cloud servers using Docker containers. The user interface and user operations of the scoring system are described in the supplementary materials.

\begin{figure}[htbp]
  \centering
    \includegraphics[width=0.8\linewidth]{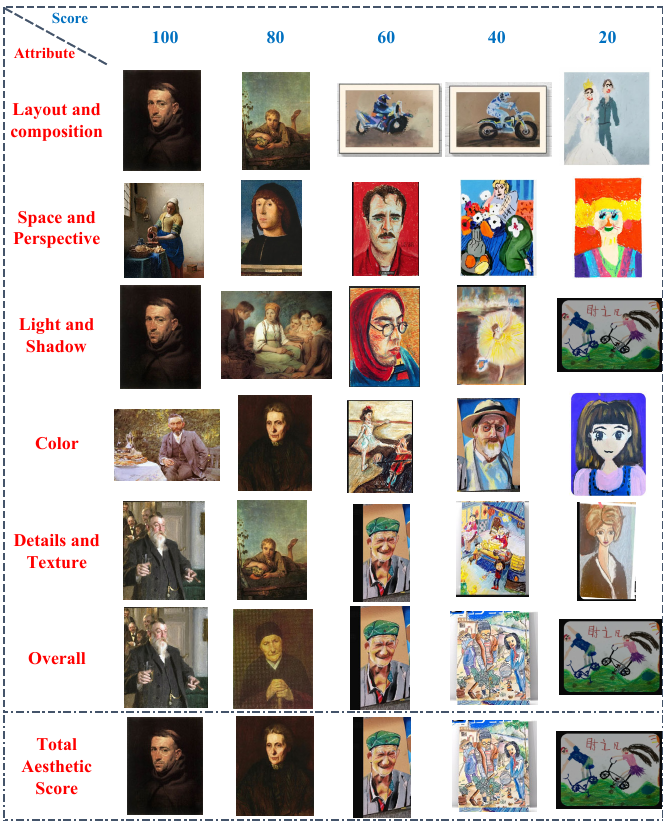}
   \caption{Representative Images for the category of oil painting - symbolism - portraiture}
   \label{Representative Images}
\end{figure}

Recognizing the advantages of professional artists in terms of experience and objectivity, we divided the scoring process into two phases. The first phase involved 27 professional artists from the oil painting, sketching, and Chinese painting groups. Their task was to lay the foundation for the entire scoring activity. During this phase, they collaboratively established scoring criteria for each image category and aesthetic attribute, presenting them in the form of benchmark images, as shown in Figure \ref{Representative Images}, illustrating a representative image example for the oil painting - symbolism - landscape category. Based on the selected benchmark images, the scoring team developed a consistent and objective scoring system, ensuring subsequent scoring work could adhere to unified standards. The first phase spanned 15 days, providing a solid theoretical foundation for the scoring activity.

The second phase involved 24 students with high academic qualifications from the oil painting, sketching, and Chinese painting groups. Leveraging the insights gained from the first phase, the second phase progressed more rapidly, completing in a total of 7 days. The division into these two phases not only maximized the use of professional artists' experience but also allowed the student scoring team to efficiently complete the scoring task under the standards and guidance accumulated in the earlier phase.

In the end, we collected a total of 31,100 annotation records from 51 professional annotators, averaging 6.24 annotations per image. Each annotation record included the total aesthetic scores and aesthetic attribute scores of the corresponding painting work. Throughout the annotation process, we meticulously assigned tasks to each scorer, specifying the art categories and the number of images involved, ensuring that each image in the APDD dataset was evaluated by at least 6 individuals. Through a comprehensive assessment of all scores, we calculated average scores based on the evaluations of all annotators, ultimately gathering the final total scores and attribute scores for each painting work.

\begin{figure}[htbp]
  \centering
    \includegraphics[width=1\linewidth]{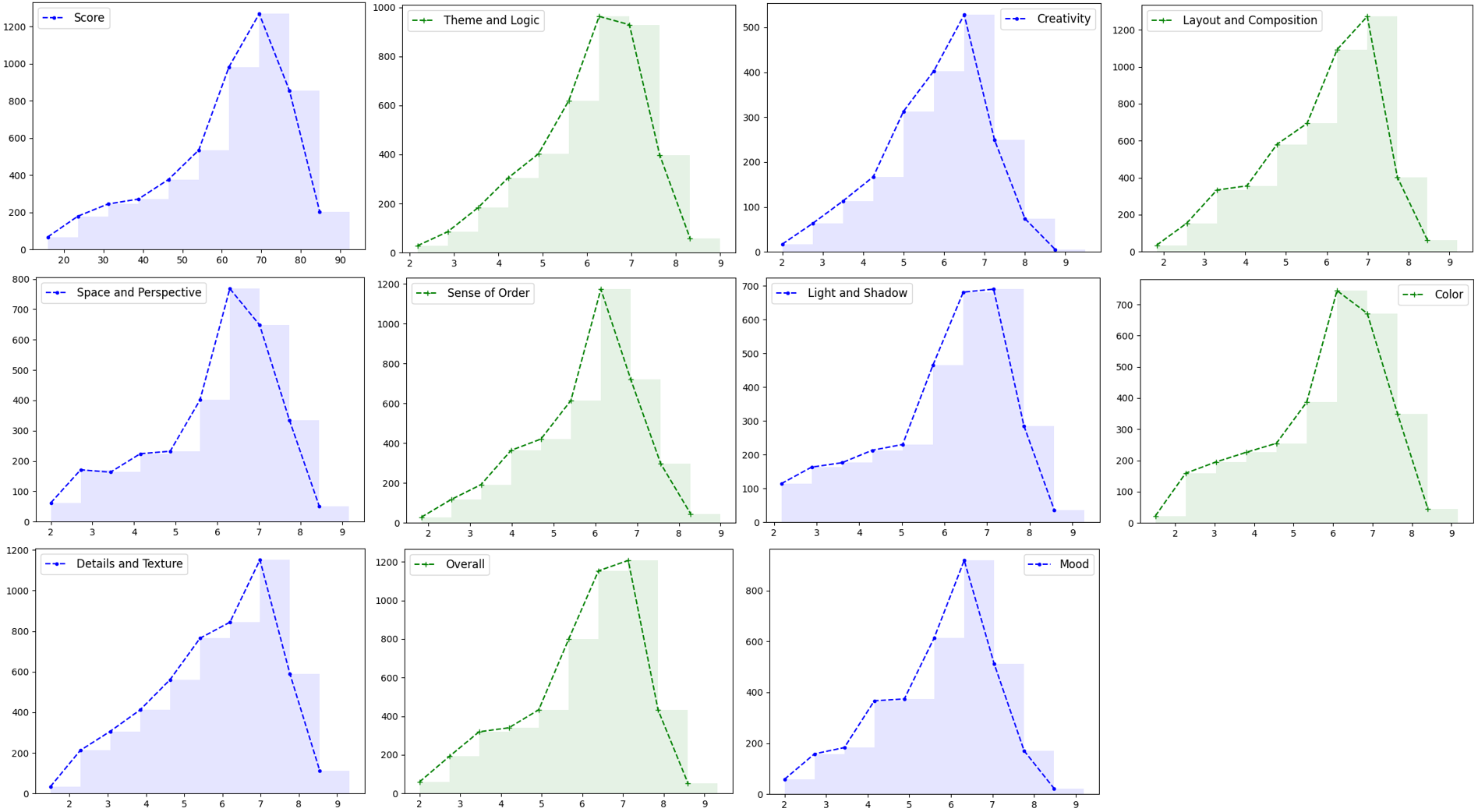}
   \caption{The score distribution of APDD. This distribution suggests a consensus on the aesthetic appeal of images, with most perceived as having a moderate level of appeal, while extreme scores are less frequent.}
   \label{Histogram}
\end{figure}

\section{Art Assessment Network for Specific Painting Styles}

\subsection{Network Architecture}

The entire network consists of a backbone network and eleven scoring branch networks, which include one total aesthetic scores evaluation network and ten aesthetic attribute score evaluation networks.

\begin{figure}[htbp]
  \centering
    \includegraphics[width=1\linewidth]{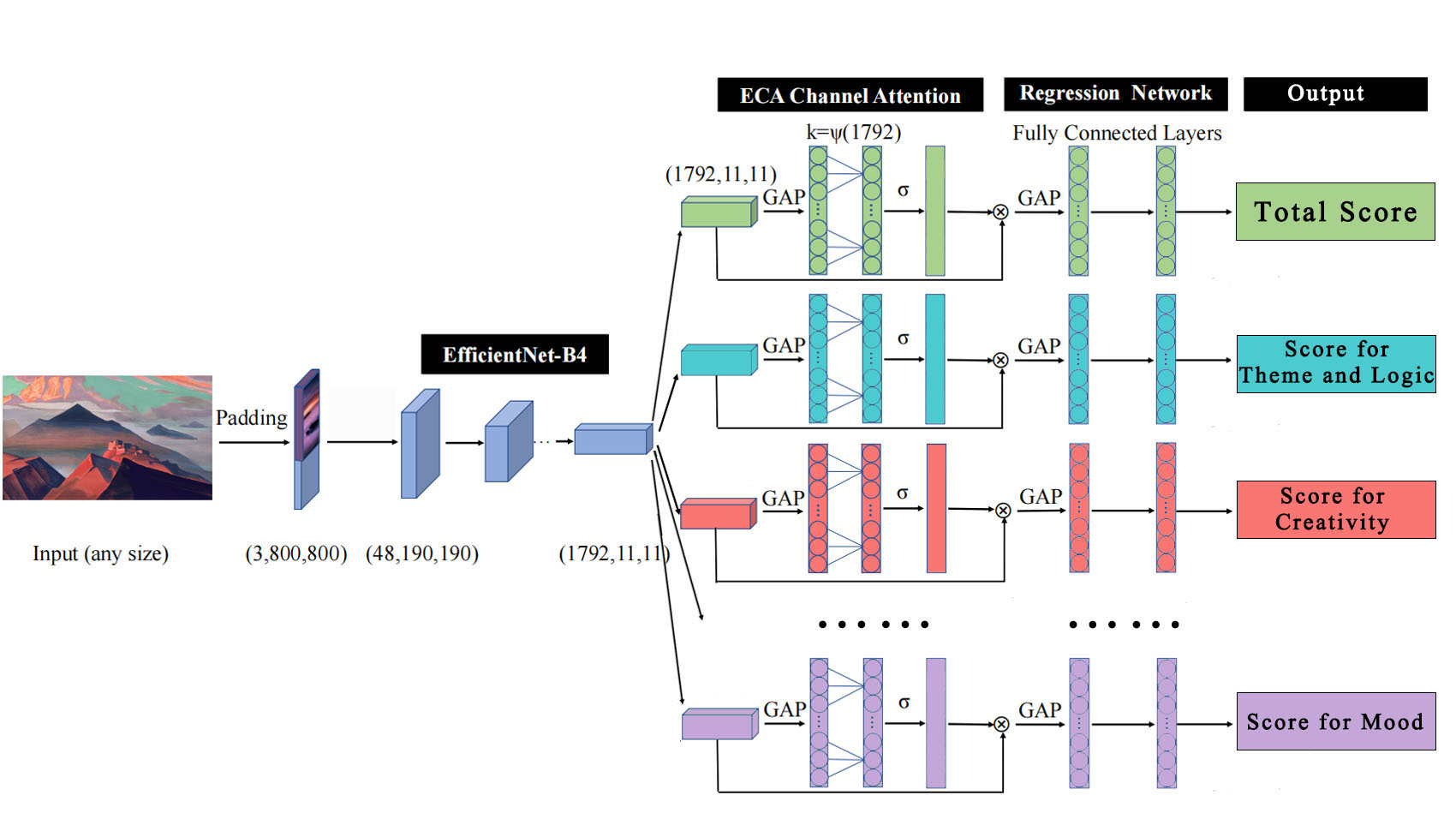}
   \caption{Overall architecture of the proposed AANSPS.}
   \label{AANSPS}
\end{figure}

The backbone network extracts features from input images to obtain input feature vectors. The specific process is as follows: First, we resize the longer side of the image proportionally to 800 pixels and pad the shorter side with zeros to achieve a size of 800 pixels. Next, within the initial convolutional operation of EfficientNet-B4, the 48×190×190 feature vector is reshaped using an adaptive average pooling layer, with 48 representing the number of feature channels. Lastly, the transformed feature vector undergoes further processing through the subsequent layers of the EfficientNet-B4 network, yielding an input feature vector of dimensions 1792×11×11.

The scoring branch network comprises a channel attention module and regression network. In order to reduce the model's complexity, this paper cites the Efficient Channel Attention (ECA) module ~\cite{wang2020eca}. It employs a local cross-channel attention mechanism and adaptive 1D convolutional kernels, which can significantly enhance performance. The ECA module aggregates the input feature vector using global average pooling (GAP), resulting in 1D vector where each channel's value represents the global average of features in that channel. Subsequently, the ECA module conducts rapid 1D convolution operations on this vector with the kernel size of \( k \), which dynamically adapts to the channel dimension \( C \) to accommodate varying channel counts. Following this, the resulting channel weights undergo Sigmoid activation, transforming them into values ranging from 0 to 1, denoting the importance or activation level of each channel. These activated channel weights are then utilized to fuse with the input feature vector, ultimately producing the fused feature vector.

 \begin{table*}[htbp]
		
		\centering
  \resizebox{1.0\linewidth}{!}{%
		\begin{tabular}{c|ccc|ccc|cccc}
   
			\noalign{\smallskip}\hline\noalign{\smallskip}

                \textbf{\multirow{2}{*}{Score Type}} &  & \textbf{MTCNN} &  &  & \textbf{DCNN} &  & &\textbf{AANSPS} & \\
                
			 & \textbf{MSE $\downarrow$ } & \textbf{MAE $\downarrow$ } & \textbf{SROCC $\uparrow$ } & \textbf{MSE $\downarrow$ } & \textbf{MAE $\downarrow$ } & \textbf{SROCC $\uparrow$ } & \textbf{MSE $\downarrow$ } & \textbf{MAE $\downarrow$ } & \textbf{SROCC $\uparrow$ }\\
			
			\noalign{\smallskip}\hline\noalign{\smallskip}
			 Total Aesthetic Score & 0.0248 & 0.1230 & 0.0561 & 0.2240 & 0.1147 & -0.1309 & \textbf{0.0149} & \textbf{0.0891} & \textbf{0.6085} \\
			
			\noalign{\smallskip}
			 Theme and Logic & 0.0138 & 0.0919 & -0.0264 & \textbf{0.0135} & \textbf{0.0914} & 0.0215 & 0.0143 & 0.0938 & \textbf{0.3896}\\
   
            \noalign{\smallskip}
			 Creativity & \textbf{0.0153} & 0.1000 & -0.0419 & 0.0154 & \textbf{0.0994} & 0.0410 & 0.0164 & 0.0996 & \textbf{0.4624}\\

            \noalign{\smallskip}
			 Layout and Composition & 0.0188 & 0.1116 & 0.0453 & \textbf{0.0174} & \textbf{0.1052} & 0.0947 & 0.0505 & 0.1139 & \textbf{0.4508} \\

             \noalign{\smallskip}
			 Space and Perspective & 0.0235 & 0.1266 & -0.1146 & 0.0198 & 0.1126 & 0.0628 & \textbf{0.0109} & \textbf{0.0828} & \textbf{0.5288}\\
   
             \noalign{\smallskip}
			 Sense of Order & 0.0163 & 0.1023 & -0.0864 & 0.0159 & 0.1011 & -0.0212 & \textbf{0.0111} & \textbf{0.0851} & \textbf{0.5433}\\

            \noalign{\smallskip}
			 Light and Shadow & 0.0238 & 0.1249 & -0.1038 & 0.0189 & 0.1070  &  -0.0273 & \textbf{0.0118} &  \textbf{0.0826} & \textbf{0.6721}\\
   
             \noalign{\smallskip}
			 Color & 0.02373 & 0.1256 & -0.0158 & 0.0212 & 0.1198 & 0.1731 & \textbf{0.0139} & \textbf{0.0890} & \textbf{0.4816}\\
   
            \noalign{\smallskip}
			 Details and Texture & 0.0247 & 0.1285 & 0.0110 & 0.0216 & 0.1208 & -0.0636 & \textbf{0.0139} & \textbf{0.0890} & \textbf{0.5381} \\

            \noalign{\smallskip}
			 Overall & 0.0203 & 0.1141 & -0.0319 & 0.0168 & 0.1036 & -0.0102  & \textbf{0.0122} & \textbf{0.0862} & \textbf{0.4961}\\

            \noalign{\smallskip}
			 Mood & 0.0194 & 0.1133 & -0.4448 & 0.0184  & 0.1096 & 0.0578 &  \textbf{0.0103} & \textbf{0.0818} & \textbf{0.6284}\\
   
			\noalign{\smallskip}\hline
		\end{tabular}
    }
    \caption{Comparison of MTCNN, DCNN and AANSPS on APDD.}
    \label{tab: Comparison of MTCNN and AANS}
	\end{table*}

The specific size \( k \) mentioned above is formulated
as follows:

$$ k=\psi(c)=\left | \frac{\log_{2}{C} }{\gamma } +\frac{b}{\gamma }  \right |_{odd}  $$

\(\left |  \right | _{odd} \) indicates the operation of rounding to the nearest odd number. The value of \( \gamma \) is 2, and the value of \( b \) is 1. We can get \( k \) = 7 if \( C \) = 1792.

The Regression Network consists of a GAP and three linear layers. Specifically, the fused feature vector is input into the GAP to obtain dimensionality-reduced feature. Subsequently, the dimensionality-reduced feature is flattened into a 1D vector. This 1D vector is then fed into three linear layers for linear mapping and non-linear transformation, ultimately yielding a corresponding score.

The loss function for the Regression Network is formulated as follows:
$$ Loss_{regre} = \frac{1}{N}\sum_{i}(s_i-\hat{s_i})^2 $$

$ N $ is the number of samples. $ s_i $ is the output score, and $ \hat{s_i} $ is the ground-truth score.

\subsection{Training Process}

We set the batch size to 64, with a learning rate of 0.0001. Adam optimizer is employed, with beta parameters set to (0.98, 0.999), and weight decay set to 0.0001. If the the regression loss does not decrease for two consecutive rounds, the learning rate is multiplied by 0.5. Our experiments were conducted using PyTorch 1.5.0 and Nvidia TITAN XPs. The dataset used in this study is APDD, which is divided into two parts: the training set and the validation set in a 9:1 ratio. We utilized the ~\cite{jin2022pseudo} model as the pretraining model. 

During training, we load the pretraining model into the total aesthetic score branch network and train this branch network based on the training set of APDD to obtain the first scoring model. Then, we use the first scoring model as the pretrained model and individually train the scoring branch networks for each attribute. When training any attribute branch network, the parameters of other scoring branch networks need to be frozen. After each branch network is trained, it will include the previously trained attribute branch networks. Once all attribute branch networks have been trained, we obtain the final scoring model.

\section{Experiments}

For comparative experiments, we selected MTCNN ~\cite{leida2022} and DCNN \cite{malu2017learning}, which are capable of simultaneously outputting both aesthetic total scores and aesthetic attribute scores. In the case of MTCNN and DCNN, we duplicated its attribute branches for a total of 10 branches. Following the same training protocol as AANSPS, we used the original model as the pretraining model. After training each branch, the newly generated model parameters from that branch were used to update the pretraining model. 

We utilize Mean Squared Error (MSE), Mean Absolute Error (MAE), and Spearman's Rank Order Correlation Coefficient (SROCC) to evaluate performance. Table \ref{tab: Comparison of MTCNN and AANS} presents the comparison of the three models in terms of MSE, MAE, and SROCC.

\begin{figure}[htbp]
  \centering
    \includegraphics[width=\linewidth]{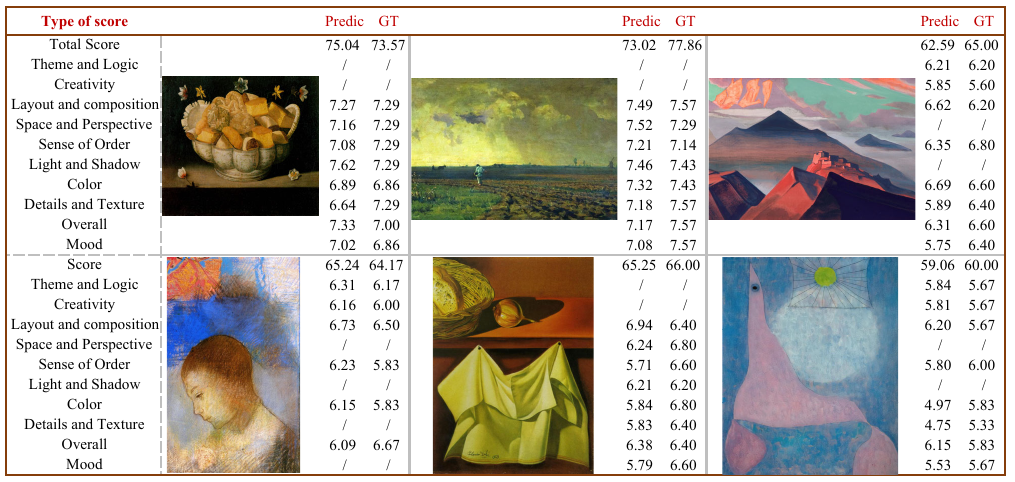}
   \caption{Test samples. "Predic" represents the predicted score of the AANSPS output. "GT" represents the ground-truth score.}
   \label{fig:onecol}
\end{figure}

\section{Conclusion}

The construction of a multi-attribute, multi-category dataset in the field of painting aesthetics represents a pioneering new task. Through close collaboration with approximately 60 global professional artists and students with high academic qualifications, we have successfully established a clear system for considering the aesthetic components of art images. We initially categorized the field of painting into 24 artistic categories and 10 aesthetic attributes, meticulously defining scoring criteria for corresponding attributes within different artistic categories. We have created the APDD dataset, the most broadly categorized and attribute-rich dataset in the field of painting, comprising 4,985 images of various painting types, accompanied by over 31,100 annotations for both total aesthetic scores and aesthetic attribute scores.Furthermore, based on the APDD dataset, we introduced the AANSPS painting image evaluation network, demonstrating its effectiveness in assessing total and attribute scores for artistic images.

However, given the vast scope of painting as an art field, the categorizations and attributes proposed in our study remain relatively limited. Future work necessitates continuous expansion of aesthetic categorizations and attributes to more comprehensively evaluate the aesthetic quality of paintings. Additionally, we plan to increase the number of images in the APDD dataset, providing score annotations for more aesthetic attributes, and supplementing these with more detailed language comments to enhance the interpretability of the scoring results.

\bibliographystyle{named}
\bibliography{ijcai24}

\section*{Appendix A}

Table \ref{tab: 4} lists the information of the 15 experts involved in annotating the APDD dataset. Due to some individuals' reluctance to disclose specific personal information, "-" is used in the table to omit the undisclosed details. We will endeavor to obtain consent from the additional 13 experts to complete the information in the table.

\begin{table*}[t]
		
		\centering

  \resizebox{1.0\linewidth}{!}{%
		\begin{tabular}{cccccccc}
   
			\noalign{\smallskip}\hline\noalign{\smallskip}
			  \textbf{Name} & \textbf{Graduate institutions} & \textbf{Degree} & \textbf{Affiliation}\\
			
			\noalign{\smallskip}\hline\noalign{\smallskip}
			Yi Lu & Central Academy of Fine Arts & Bachelor & Central Academy of Fine Arts   \\
			Shan Gao  & Beijing University of Technology & Master & Beijing Tianqi Century Culture Communication Co.    \\
   			Guangdong Li  & Beijing Institute of Fashion Technology & Master & 51meishu    \\
         	Mianhong Han  & Beijing Institute of Graphic Communication & - & -    \\
                Tingting Ma  & Beijing Institute of Fashion Technology & Master & Taiyuan Normal University    \\
                Rui Li  & Central Academy of Fine Arts & Master & China Academy of Building Research Co. Ltd.    \\
                Jialun Cao  & College of Design, Wuxi University of Light Industry & Bachelor & -    \\
                Jie Wang  & Academy of Fine Arts, Munich & Master & -    \\
                Keyao Zhu  & Central Academy of Fine Arts & Bachelor & Central Academy of Fine Arts    \\
                Xingming Liu  & - & Bachelor & -    \\
                Qing Cai  & Central Academy of Fine Arts & Master & Central Academy of Fine Arts    \\
                Yi Zhang  & Central Academy of Fine Arts & Bachelor & Central Academy of Fine Arts    \\
                Jianfei Liu  & Central Academy of Fine Arts & Bachelor & -    \\
                Sizhe Fan  & Communication University of China & - & Beijing Xilin Calligraphy and Painting Academy    \\
                Wei Liu  & College of Fine Arts, Peking University & Master & -    \\

			\noalign{\smallskip}\hline
		\end{tabular}
  }
    \caption{Information for some experts.}
    \label{tab: 4}
	\end{table*}

\end{document}